\documentclass[10pt, conference, compsocconf]{IEEEtran}
\ifCLASSINFOpdf
  \usepackage[pdftex]{graphicx}
  \DeclareGraphicsExtensions{.pdf,.jpeg,.png}
\else
\fi
\usepackage{lipsum}
\usepackage{xcolor}
\usepackage{tikz}
\usetikzlibrary{positioning}
\usepackage{amsmath} % assumes amsmath package installed
\usepackage{amssymb}  % assumes amsmath package installed
\usepackage{bm}
\usepackage{siunitx}
\usepackage{tabularx,ragged2e,booktabs}
\newcolumntype{C}{>{\Centering\arraybackslash}X}
\usepackage{hyperref}
\begin{document}
%
% paper title
% can use linebreaks \\ within to get better formatting as desired
\title{Continuous-Time Range-Only Pose Estimation}

% author names and affiliations
% use a multiple column layout for up to two different
% affiliations
\author{
	\IEEEauthorblockN{Abhishek Goudar\IEEEauthorrefmark{1}\IEEEauthorrefmark{2}, Timothy D. Barfoot\IEEEauthorrefmark{1}\IEEEauthorrefmark{2}, Angela P. Schoellig\IEEEauthorrefmark{1}\IEEEauthorrefmark{2}\IEEEauthorrefmark{3}}
	\IEEEauthorblockA{\IEEEauthorrefmark{1} University of Toronto Institute for Aerospace Studies (UTIAS), Toronto, Canada}
	\IEEEauthorblockA{\IEEEauthorrefmark{2}Vector Institute for Artificial Intelligence, Toronto, Canada}
	\IEEEauthorblockA{\IEEEauthorrefmark{3} Munich Institute for Robotics and Machine Intelligence (MIRMI), Technical University of Munich, Munich,  Germany \\
abhishek.goudar@robotics.utias.utoronto.ca, tim.barfoot@utoronto.ca, angela.schoellig@tum.de }
	
}
\maketitle

\begin{abstract}

Range-only (RO) localization involves determining the position of a mobile robot by measuring the distance to specific anchors. RO localization is challenging since the measurements are low-dimensional and a single range sensor does not have enough information to estimate the full pose of the robot. As such, range sensors are typically coupled with other sensing modalities such as wheel encoders or inertial measurement units (IMUs) to estimate the full pose. In this work, we propose a continuous-time Gaussian process (GP)-based trajectory estimation method to estimate the full pose of a robot using only range measurements from multiple range sensors. Results from simulation and real experiments show that our proposed method, using off-the-shelf range sensors, is able to achieve comparable performance and in some cases outperform alternative state-of-the-art sensor-fusion methods that use additional sensing modalities.

\end{abstract}


% For peer review papers, you can put extra information on the cover
% page as needed:
% \ifCLASSOPTIONpeerreview
% \begin{center} \bfseries EDICS Category: 3-BBND \end{center}
% \fi
%
% For peerreview papers, this IEEEtran command inserts a page break and
% creates the second title. It will be ignored for other modes.
\IEEEpeerreviewmaketitle

\newcommand{\state}{\mathbf{T}^\mathcal{W}_{i}}

\newcommand{\pose}{\mathbf{T}}
\newcommand{\nompose}{\bar{\mathbf{T}}}
\newcommand{\perturb}{\bm{\xi}}
\newcommand{\world}{\mathcal{W}}

\section{Introduction}

Accurate localization is essential for the reliable operation of any autonomous system. Generally, different sensing modalities are used for localization in different environments. In outdoor environments, the Global Positioning System (GPS) \cite{kaplan} is one of the preferred methods of localization. In recent years, localization against a map using lidars and cameras has also been widely adopted in self-driving cars.  For indoor environments, the sensing modalities include vision, laser, magnetic tapes, and radio wave-based sensors such as WiFi and ultrawideband (UWB) \cite{raza}. 

\begin{figure}[t]
	\vspace*{1em}
	\centering
	\includegraphics[scale=0.33]{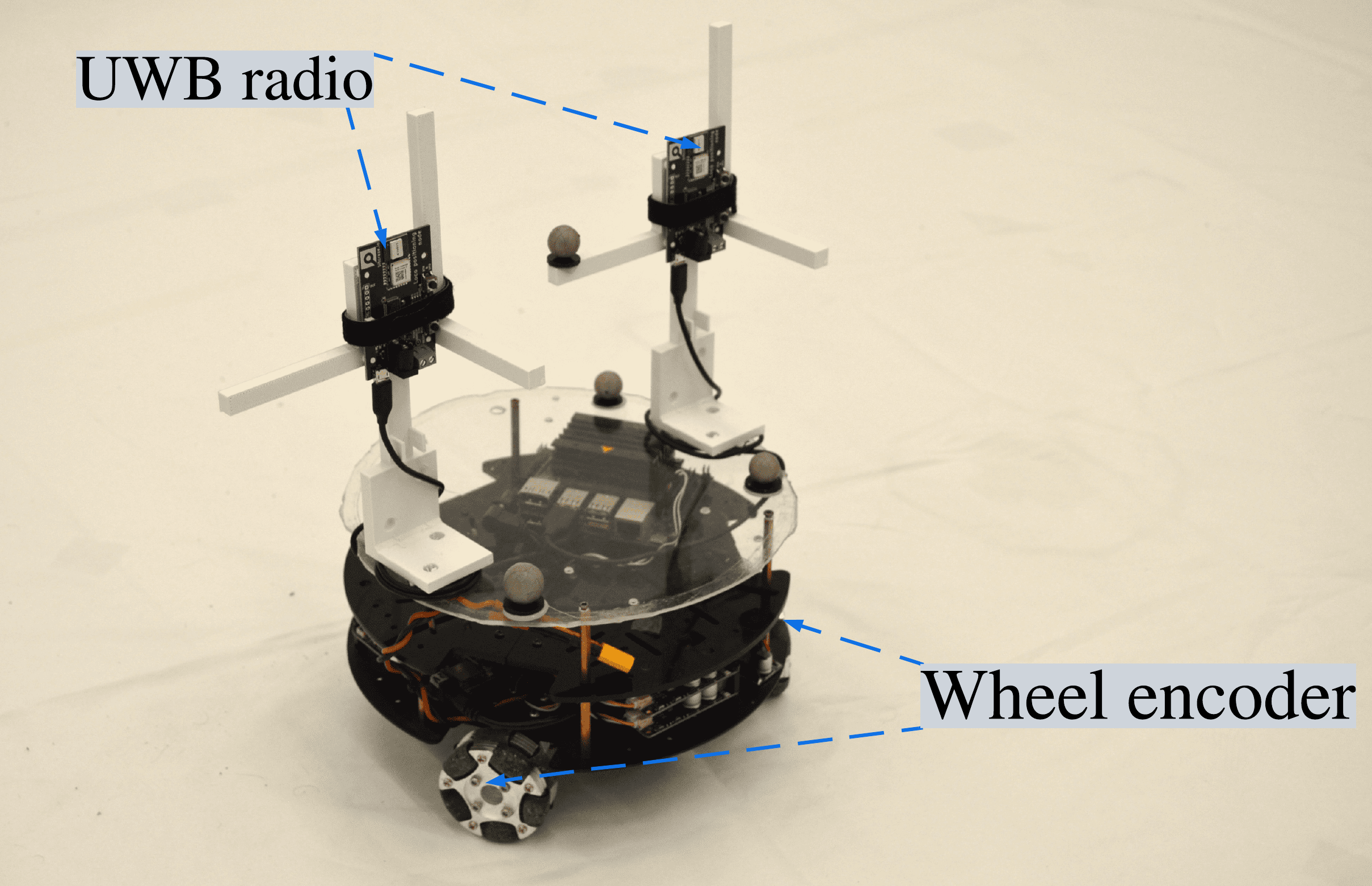}
	\caption{Our test platform for 2D trajectory estimation is a custom-built wheeled holonomic robot. It is equipped with two ultrawideband (UWB) radios to estimate the full 2D pose using continuous-time trajectory estimation. The robot also has 3 mecanum wheels with encoders, which are used for comparison with the baseline algorithm.}
	\label{fig:omniloc}
\end{figure}

The underlying principle of radio wave-based positioning technologies such as GPS and UWB is point-to-point range measurements between a transmitter and a receiver. In range-only (RO) localization, a robot with a range sensor such as a radio measures its distance to other radios known as \textit{anchors}. The distance measurements are then combined to determine the position of the robot \cite{kaplan}. RO localization is challenging since range measurements are sparse and hence are typically used only for position estimation. In the case where the full pose of the robot is needed, range sensors are typically used in combination with additional sensors as wheel encoders \cite{Blanco} or inertial measurement units (IMUs) \cite{hol2009}. A common drawback of such sensor-fusion methods is that sufficient excitation or movement is required before the full pose can be determined unambiguously \cite{trawny2010,goudar2021}. For example, in tightly-coupled UWB-IMU systems, excitation of the accelerometer and gyroscope axes is necessary for full-state observability \cite{goudar2021}. This can be a limiting factor for autonomous robots that frequently encounter \textit{stop-and-go} motion patterns in settings such as warehouses and factories. Additionally, sensors such as wheel encoders are susceptible to wheel slippage, which can result in poor performance, especially in slippery and off-road conditions. 
 
In this work, we propose a continuous-time trajectory estimation method that is able to estimate the full pose of a robot using only range measurements from multiple range sensors. Unlike conventional multimodal sensor-fusion algorithms, the proposed method does not require excitation or motion for full pose estimation, although it may still benefit from it. Additionally, the proposed method is not affected by naturally occurring conditions such as wheel slippage and lack of adequate motion.

In summary, the main contributions of this work are \textit{(i)} a continuous-time approach to 2D and 3D pose estimation using only range measurements from multiple range sensors, and \textit{(ii)} demonstration of the proposed approach in simulation and real experiments.

The paper is organized as follows. We review the related work in Section \ref{sec:related_work}. We formulate our problem in Section \ref{sec:prob_statement} and describe our proposed method in Section \ref{sec:methodology}. We present results from evaluation of our approach in simulation and real experiments in Section \ref{sec:experiments}. We include a discussion of the results and challenges of our approach in Section \ref{sec:discussion} and conclude the paper in Section \ref{sec:conclusion}.
\section{Related work} \label{sec:related_work}
Range-only (RO) localization has a rich history since it is widely used in positioning technologies such as Global Position System (GPS) \cite{kaplan}, and more recently in other wireless positioning technologies such as ultrawideband (UWB) \cite{yavari2014}. 

As alluded to previously, the sparse nature of range measurements does not afford full pose estimation. As such, range sensors are typically combined with wheel odometry \cite{Djugash,Blanco,zhouwheel} for the localization of ground robots. Another common approach to pose estimation is to combine range sensors with inertial measurement units (IMUs) \cite{hol2009,Mueller2015,goudar2021}. More recently, other sources of pose such as fiducial markers \cite{Hoeller2017} and visual-inertial-odometry have been used as well \cite{Nguyen2021, cioffi2020tightly}. Common estimation frameworks used for positioning include parametric filtering methods, \cite{hol2009,Mueller2015, goudar2021}, nonparametric filtering methods \cite{Blanco}, and more recently, optimization-based methods \cite{Nguyen2021} have gained traction. Most of the previous works use a discrete-time formulation for trajectory estimation. 

In recent years, there has been an interest in continuous-time approach to RO localization. A spline-based approach to the fusion of UWB and IMU data for continuous-time trajectory estimation is proposed in \cite{splineimu}. In \cite{kiki2019}, polynomial basis functions are used to parameterize the robot trajectory and to derive the conditions necessary for recovering the trajectory. More recently, continuous-time trajectory estimation based on Gaussian process (GP) regression \cite{Barfoot2014a} was applied to RO localization \cite{kiki2022}. Unlike the previous works, \cite{kiki2019, kiki2022} use only range measurements to estimate the robot position over time. The application of range measurements for 2D relative pose estimation between multiple agents using multiple range sensors was recently shown in \cite{fishberg}.

An alternative approach to pose estimation using range measurements from multiple range sensors is to combine multilateration \cite{kaplan} with 3D point set registration \cite{arunsmethod}. A limitation of this approach is that it requires synchronization between the anchors and the range sensors. 

In this work, we take the GP regression approach to continuous-time trajectory estimation of \cite{Barfoot2014a,Anderson2015} and perform full 2D and 3D pose estimation using only range measurements from multiple range sensors. In contrast to previous methods, our method does not require additional sensing modalities nor does it require synchronization between the anchors and the range sensors. To the best of the authors' knowledge, continuous-time 2D and 3D pose estimation using only asynchronous range measurements from multiple range sensors has not appeared in the literature.

\section{Problem statement} \label{sec:prob_statement}
The setup we consider is that of a robot navigating in an environment where multiple anchors have been installed. The objective of this work is to estimate the pose of the robot using only range measurements made to the anchors. We assume that
\begin{itemize}
	\item multiple ($\geq 3$) non-collated anchors are installed in the environment,
	\item the position of the anchors is known,
	\item each robot is equipped with at least 2 non-collocated range sensors for 2D pose estimation and at least 3 range sensors for 3D pose estimation, and
	\item the position of the range sensors in the robot body frame is known.
\end{itemize}

The proposed method does not require range measurements between the anchors and the range sensors to arrive simultaneously in a synchronized manner. Specifically, a single range measurement between a range sensor on the robot and an anchor is necessary at each time step.

\section{Methodology} \label{sec:methodology}

\begin{figure}[!t]
	\center
	\begin{tikzpicture}[
	node distance={10mm},
	vertex/.style={circle, draw=black!100, thick, minimum width=0.9cm},
	empty_factor/.style={minimum size=1mm},]
	% nodes
	\node[circle,fill=black,inner sep=0pt,minimum size=5pt] (prior) at (0,0) {};
	\node[vertex] (x0) [right of=prior] {$\mathbf{x}(t_0)$};
	\node[circle, draw, right of=x0, fill=black, inner sep=0pt,minimum size=5pt, label=below:{$\mathbf{o}(t_0)$}] (o0){};
	\node[vertex] (x1)  [right of=o0] {$\mathbf{x}(t_1)$};
	\node[circle, draw, right of=x1, fill=black, inner sep=0pt,minimum size=5pt, label=below:{$\mathbf{o}(t_1)$}] (o1){};
	\node[vertex] (x2)  [right of=o1] {$\mathbf{x}(t_2)$};
	\node[empty_factor] (o2) [right of=x2] {$\hdots$};
	\node[vertex] (xM) [right of=o2] {$\mathbf{x}(t_k)$};
	\node[circle, draw, above of=x1, fill=black, inner sep=0pt,minimum size=5pt, label=left:{$r_1(t_1)$}] (y1){};
	\node[circle, draw, above of=o1, fill=black, inner sep=0pt,minimum size=5pt, label=right:{$r_1(t_2)$}] (y2){};
	\node[circle, draw, above of=xM, fill=black, inner sep=0pt,minimum size=5pt, label=right:{$r_j(t_k)$}] (yM){};
	\node[vertex] (xa) [above of=y1, fill=gray, opacity=0.5, text opacity=1] {$\mathbf{p}_{a_1}$};
	\node[vertex] (xb) [above of=yM, fill=gray, opacity=0.5, text opacity=1] {$\mathbf{p}_{a_j}$};
	
	%
	% edges
	\draw (prior) -- (x0);	
	\draw (x0) -- (o0);
	\draw (o0) -- (x1);
	\draw (x1) -- (o1);
	\draw (o1) -- (x2);
	\draw (x2) -- (o2);
	\draw (o2) -- (xM);
	\draw (x1) -- (y1);
	\draw (y1) -- (xa);
	\draw (x2) -- (y2);
	\draw (y2) -- (xa);
	\draw (xM) -- (yM);
	\draw (yM) -- (xb);
	%\draw [thick, dash pattern=on 1pt off 9pt] (xa) -- (xb);
	%\draw [thick, dash pattern=on 1pt off 9pt] (y2) -- (yM);
	%
	\end{tikzpicture}
	\caption{Factor graph for a range-only (RO) localization setup. In RO localization, a robot equipped with a range sensor, such as a wireless radio, estimates its position by measuring the distance to other wireless radios, known as \textit{anchors}, installed in the environment. The trajectory consists of a set of nodes representing the robot state, $\mathbf{x}(t)$. The anchor positions, $\mathbf{p}_{a_\#}$, are known, as indicated by the filled circles. Motion prior factors are denoted by $\mathbf{o}(t)$ and the range measurements by factors $r_{\#}(t)$, where $\#$ denotes the anchor id.}
	\label{fig:factor_graph}
	\vspace{-1em}
\end{figure}
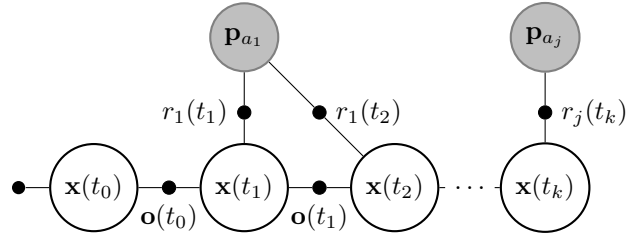

\subsection{Preliminaries}

We introduce the frame convention and the notation that will be used throughout the paper. We denote the world frame by $\mathcal{F}_\mathcal{W}$ and the robot frame by $\mathcal{F}_i$. We represent the robot pose in the world frame using elements of the special Euclidean \textit{Lie} group $\mathbf{T}_{wi} \in {SE}(n)$, where $n=2$ for 2D pose and $n=3$ for 3D pose. We will use 3D poses for exposition with the understanding that the proposed method carries over to 2D poses. A generic pose element $\mathbf{T}$ is parameterized as $\mathbf{T} = \{ \mathbf{p}, \mathbf{R} \}$, where $\mathbf{p} \in \mathbb{R}^{3 \times 1} $ represents the position and $\mathbf{R} \in SO(3)$, a member of the special orthogonal group, represents the orientation. We use the \textit{right perturbation} convention of \cite{Barfoot2023} to represent perturbations around the nominal pose. Specifically, a generic pose is decomposed into a nominal pose, $\bar{\mathbf{T}} \in SE(3)$, and a small perturbation $\bm{\xi} \in \mathbb{R}^{6 \times 1}$ as
\begin{equation}
\mathbf{T} = \bar{\mathbf{T}} \exp(\bm{\xi}^\wedge),
\label{eqn:pose_rep}
\end{equation}
where the operator, $(\cdot)^\wedge$, maps an element of $\mathbb{R}^{6 \times 1}$ to an element of the \textit{Lie algebra}, $\frak{se}(3)$. The $\exp(\cdot)$ operator is a \textit{retraction} operation for $SE(3)$ and maps an element of the Lie algebra $\frak{se}(3)$ back to the Lie group, $SE(3)$. 

%We represent uncertainty on poses using the formulation of \cite{Barfoot2014} but with the \textit{right perturbation} convention \cite{Sola2018}. Specifically, a generic pose is decomposed into a nominal pose $\bar{\mathbf{T}} \in SE(3)$ and a small perturbation $\bm{\xi} \in \mathbb{R}^{6 \times 1}$:
%\begin{equation}
%\mathbf{T} = \bar{\mathbf{T}} \exp(\bm{\xi}^\wedge),
%\label{eqn:pose_rep}
%\end{equation}
%where $\exp$ is the \textit{retraction} operation for $SE(3)$, and $(\cdot)^\wedge$ is the \textit{hat} operator that maps an element of $\mathbb{R}^6$ to an element of the \textit{Lie algebra} $\mathfrak{s}e(3)$, and $\perturb \sim \mathcal{N}(\mathbf{0}, \mathbf{\Sigma})$ is a Gaussian random variable with covariance $\mathbf{\Sigma}$.%

As mentioned previously, the setup we consider is of a robot with multiple range sensors navigating in an environment with anchors. Range measurements between the range sensors on the robot and the anchors arrive in an asynchronous manner. Since a single range measurement is not sufficient to constrain the full state, we use motion priors as constraints between subsequent range measurements to constrain the full state. The factor graph for such a setup is shown in Figure \ref{fig:factor_graph}. The motion priors are added as \textit{binary} factors, $\mathbf{o}(t_i)$, between two consecutive robot states, $\mathbf{x}(t_i)$ and $\mathbf{x}(t_{i+1})$,  and the range measurements are added as \textit{unary} factors, $r_j(t_i)$. We perform inference on the factor graph using \textit{maximum a posteriori} (MAP) estimation. Under the Gaussian noise assumption, this is equivalent to solving a nonlinear least-squares problem. Next, we describe the motion model used to generate the motion prior and the range measurement model.

\subsection{Motion model}
We adopt the Gaussian process (GP) regression approach to continuous-time trajectory estimation of \cite{Barfoot2014a} and use the white-noise-on-acceleration (WNOA) motion model \cite{Anderson2015}. Since the pose variables are nonlinear, we use the \textit{local pose variable} formulation of \cite{Anderson2015} to define a linear time-invariant (LTI) motion model on the local pose variables, which are subsequently stitched together to generate the whole trajectory. The motivation for choosing such a motion prior is that the resulting system matrices are sparse and can be solved very efficiently. For completeness, we present here an overview of the motion prior generation for the right-perturbation scheme. A more detailed description using the left-perturbation scheme can be found in \cite{Barfoot2023}.

The robot state at any time $t$ is given by $\mathbf{x}(t) = \{\mathbf{T}(t), \bm{\varpi}(t)\} \in SE(3) \times \mathbb{R}^{6 \times 1}$, where, as before, $\mathbf{T}(t)$ is the robot pose in frame $\mathcal{F}_\mathcal{W}$ and $\bm{\varpi}$ is the generalized body-centric velocity of the robot. We drop subscripts denoting the frames to reduce clutter. We define the local pose variables as perturbations around the nominal pose as
\begin{equation}
	\mathbf{T}(t) = \mathbf{T}(t_k) \exp(\bm{\xi}_k^{\wedge}(t)),
\end{equation}
where $\bm{\xi}_k \in \mathbb{R}^{6 \times 1}$ is the local pose variable. We define a motion model on the local pose variables using the following LTI stochastic differential equation (SDE) \cite{Anderson2015,Barfoot2023}:
\begin{align*}
	\frac{d}{dt}
	\begin{bmatrix}
		\bm{\xi}_k(t) \\
		\dot{\bm{\xi}}_k(t)
	\end{bmatrix} &
	=\begin{bmatrix}
	\mathbf{0} & \mathbf{I} \\
	\mathbf{0} & \mathbf{0} \\
	\end{bmatrix}
	\underbrace{\begin{bmatrix}
	\bm{\xi}_k(t) \\
	\dot{\bm{\xi}}_k(t)
	\end{bmatrix}}_{\bm{\gamma}_k(t)} 
	+ 
	\begin{bmatrix}
	\mathbf{0} \\
	\mathbf{I}
	\end{bmatrix} \mathbf{w}_k(t),
\end{align*}
where $\mathbf{I}$ is the identity matrix of appropriate dimensions, $\mathbf{w}_k(t) \sim \mathcal{GP}(\mathbf{0}, \bm{Q}(t-t'))$ is a zero-mean GP with power spectral density matrix, $\bm{Q}$. The local pose velocity, $\dot{\bm{\xi}}_k(t)$, is related to the generalized body-centric velocity as  $ \dot{\bm{\xi}}_k(t) = \bm{\mathcal{J}}_r^{-1}(\bm{\xi}_k(t)) \bm{\varpi}(t)^{\vee}$, where $\bm{\mathcal{J}}_r$ is the right jacobian of $SE(3)$ \cite{chirikjian}, and the operator $(\cdot)^{\vee}$ maps an element of the Lie algebra $\mathfrak{se}(3)$ to $\mathbb{R}^{6 \times 1}$. The above SDE can be integrated in closed form to obtain a sparse GP prior. The mean of the prior between two time instants is
\begin{equation*}
	\bm{\gamma}_{k-1}(t_k) = \bm{\Phi}(t_k, t_{k-1}) \bm{\gamma}_{k-1}(t_{k-1}),
\end{equation*}
and the corresponding covariance is
\begin{equation*}
\check{\mathbf{P}}_k(t_k) = \bm{\Phi}(t_k, t_{k-1}) \check{\mathbf{P}}_k(t_{k-1}) \bm{\Phi}(t_k, t_{k-1})^T + \mathbf{Q}(t_k - t_{k-1}),
\end{equation*}
where the system transition matrix is
\begin{align*}
\bm{\Phi}(t_k, t_{k-1}) &= \begin{bmatrix}
\mathbf{I} & \mathbf{I} \Delta t_{k:k-1} \\
\mathbf{0} & \mathbf{I}
\end{bmatrix},
\end{align*}
and the noise between two time steps is
\begin{align*}
\mathbf{Q}(t_k - t_{k-1}) &= \begin{bmatrix}
\frac{1}{3}\Delta t_{k:k-1}^3 \bm{Q} & \frac{1}{2}\Delta t_{k:k-1}^2 \bm{Q} \\
\frac{1}{2}\Delta t_{k:k-1}^2 \bm{Q} & \Delta t_{k:k-1} \bm{Q} \\
\end{bmatrix},
\end{align*}
with $\Delta t_{k:k-1} = t_k - t_{k-1}$.

The error terms corresponding to the motion prior required for MAP estimation are given by
\begin{align*}
	\mathbf{e}_p &= \begin{bmatrix} 
	\left( \bm{\varpi}(t_{k-1}) \Delta t_{k:k-1} - \ln\left( \mathbf{T}(t_{k-1})^{-1} \mathbf{T}(t_{k})  \right)  \right)^{\vee} ,\\
	\bm{\varpi}(t_{k-1})^{\vee} - \bm{\mathcal{J}}_r^{-1}(\ln( \mathbf{T}(t_{k-1})^{-1} \mathbf{T}(t_{k}))^{\vee}) \bm{\varpi}(t_{k})^{\vee}
	\end{bmatrix},
\end{align*}
where $\ln(\cdot)$ is the inverse retraction operation and converts a member of the Lie group to its Lie algebra.  
%\begin{align*}
%	\bm{\xi}_{k-1}(t_k) &= \bm{\xi}_{k-1}(t_{k-1}) + \dot{\bm{\xi}}_{k-1}(t_{k-1}) \Delta t, \\
%	\exp(\bm{\xi}_{k-1}(t_k)) &= \exp(\bm{\xi}_{k-1}(t_{k-1}) + \dot{\bm{\xi}}_{k-1}(t_{k-1}) \Delta t ), \\
	%
	%
%	\exp(\bm{\xi}_{k-1}(t_k)) &\approx \exp(\bm{\xi}_{k-1}(t_{k-1})) \\ &  \qquad  \exp(\mathcal{J}_r(\bm{\xi}_{k-1})\dot{\bm{\xi}}_{k-1}(t_{k-1}) \Delta t), \\
	%
	%
%	\mathbf{T}(t_{k-1}) \exp(\bm{\xi}_{k-1}(t_k)) &= \mathbf{T}(t_{k-1}) \exp(\bm{\xi}_{k-1}(t_{k-1})) \\ 
%	& \qquad \exp(\bm{\varpi}_{k-1}(t_{k-1}) \Delta t), \\
	%
	%
%	\mathbf{T}(t_k) &= \mathbf{T}(t_{k-1}) \exp(\bm{\varpi}_{k-1}(t_{k-1}) \Delta t),
%\end{align*}

\subsection{Range measurement model}

The range measurement at any time $t$ between the robot and anchor $j$ is given by
\begin{equation}
r_j(t) = \| \mathbf{p}_{a_j} - \mathbf{R}(t) \mathbf{p}_{u} - \mathbf{p}(t) \|_2 + \eta_{rt},
\label{eqn:range_meas_model}
\end{equation}
where $\|\cdot\|_2$ is the $\ell^2$ norm, $\mathbf{T}(t) = \{\mathbf{p}(t), \mathbf{R}(t) \}$ is the robot pose at time $t$, $\mathbf{p}_{a_j} \in \mathbb{R}^{3 \times 1}$ is the position of anchor $j$ in world frame, and $\mathbf{p}_{u}$ is the position of the range sensor w.r.t robot body frame $\mathcal{F}_i$, a.k.a \textit{lever arm}, and $ \eta_{r}(t) \sim \mathcal{N}(0, \sigma_r^2)$ is an additive white Gaussian noise of variance, $\sigma_r^2$. 

From the measurement model \eqref{eqn:range_meas_model}, we can see that the only term influencing the orientation of the robot is the lever arm: a larger lever arm provides better orientation estimation. This is especially true in the presence of noisy range measurements. Additionally, for 3D pose estimation, multiple ($\geq 3$) noncollinear range sensors are needed in order to excite all three axes of orientation. The error term for MAP estimation corresponding to the range measurement model is
\begin{align*}
	e_r &= r_j(t) - \| \mathbf{p}_{a_j} - \mathbf{R}(t) \mathbf{p}_{u} - \mathbf{p}(t) \|_2. 
\end{align*}

\subsection{Inference} 
To keep the computational cost low, we use a \textit{fixed-lag smoother} (FLS) to combine a window of range measurements and motion priors to estimate the robot trajectory. The size of the window is parameterized by time duration, $\delta t_\textrm{fls}$. A complete description of MAP estimation done in each fixed window can be found in \cite{Barfoot2023}. States older than $\delta t_\textrm{fls}$ are marginalized out. We use the GTSAM \cite{gtsam} library to implement the FLS.

\section{Experiments} \label{sec:experiments}

In this section, we present results from simulations and real experiments to demonstrate RO pose estimation under different settings. In simulation, we evaluate the effect of the range measurement noise and the lever arm length on estimation accuracy. We evaluate the proposed approach for 2D and 3D trajectory estimation in real experiments. 

\subsection{Simulations}

\begin{figure}
	\centering
	\includegraphics[scale=0.7]{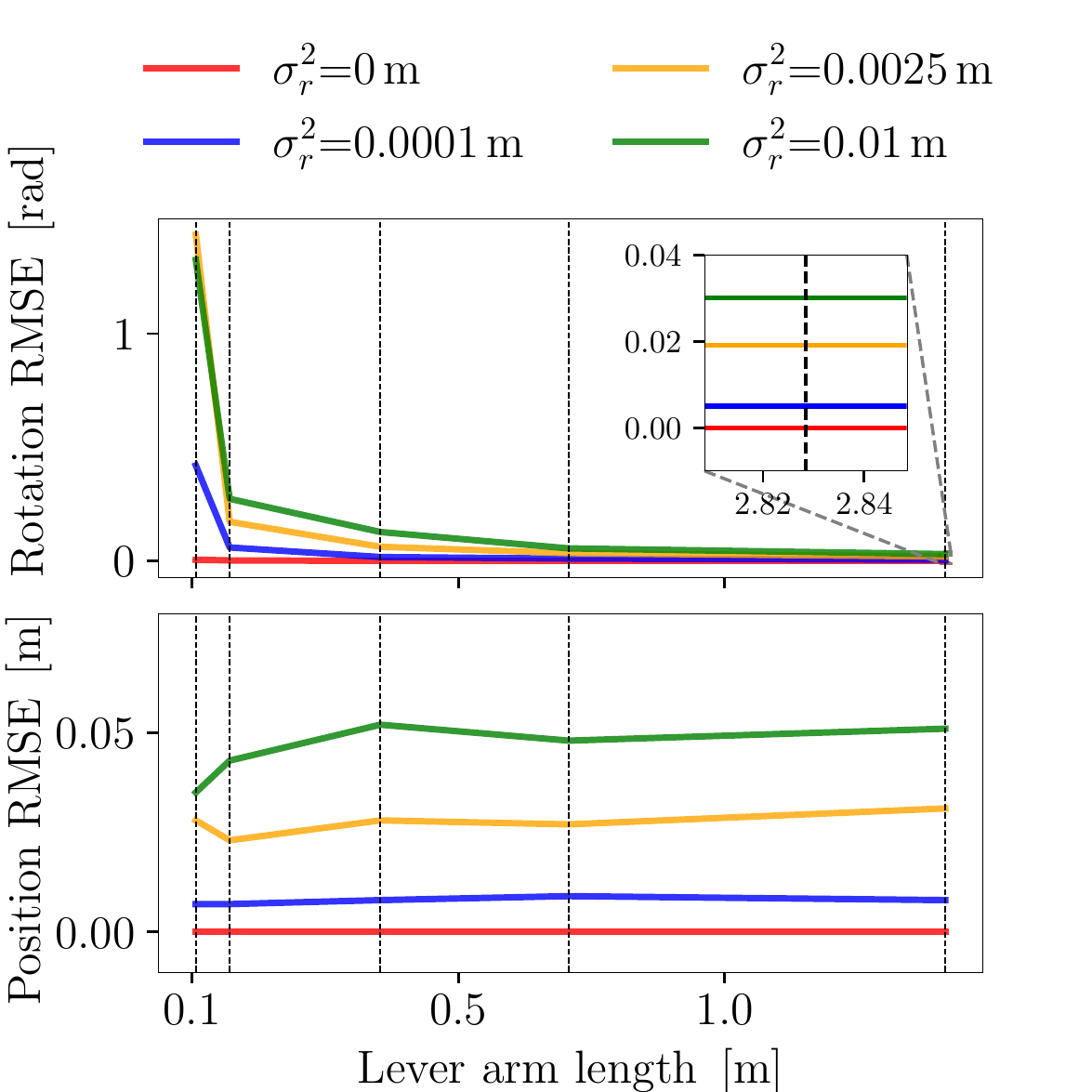}
	\caption{Results from simulations demonstrating the effect of magnitude of range measurement noise and lever arm length on orientation (top) and position (bottom) root-mean-square error (RMSE). Lever arm lengths are indicated by dashed vertical lines. For a given length of the lever arm, the range measurements are corrupted with Gaussian noise of increasing covariance (denoted by $\sigma_r^2$). The plots show that orientation estimation is more susceptible to noise in range measurements for smaller lever arms, whereas position estimation is relatively less sensitive.}
	\label{fig:sim_results}
\end{figure}

In simulations, we evaluate the sensitivity of RO localization to the magnitude of noise in range measurements and the length of the lever arm. Our simulation environment consists of 8 anchors and a quadrotor with 3 noncollinear range sensors. The quadrotor is controlled using ground truth and is commanded a straight-line trajectory. The length of the lever arm is equal for the three range sensors. The lever arm lengths are varied from $\|\mathbf{p}^i_u \|_2 = 0.014\,\si{m}$ to $\|\mathbf{p}^i_u \|_2 = 2.8\,\si{m}$. For a given length of the lever arm, the range measurements are corrupted with Gaussian noise of increasing variance ($\sigma_r^2 = 0\,\si{m}$ to $\sigma_r^2 = 0.01\,\si{m}$). Position and orientation root-mean-square error (RMSE) from multiple simulations are shown in Figure \ref{fig:sim_results}. The plots show that orientation estimation is more susceptible to noise in range measurements, especially for smaller lever arms. This is expected because the resolution of any angle relies on the physical distance between the range sensors. However, noise in range measurements can negate the effect of this physical separation.  In contrast, the estimation of position is relatively more robust to the noise in range measurements as it does not require range sensors to be physically separated. However, increasing measurement noise results in a higher position RMSE.

\subsection{Real experiments}

\subsubsection*{Setup} We use the DW1000-based \cite{dw1000} ultrawideband (UWB) radios from Bitcraze as both anchors and range sensors on the robot. The test space consists of an arena of dimensions $7\,\si{m} \times 8\,\si{m} \times 3.5\,\si{m}$ with 8 UWB anchors installed in the corners of the space. The UWB radios are operated in two-way-range (TWR) mode. The test space is also equipped with a Vicon motion capture system, which is used as a source of ground truth pose.

\begin{figure}[t]
	\includegraphics[scale=0.92]{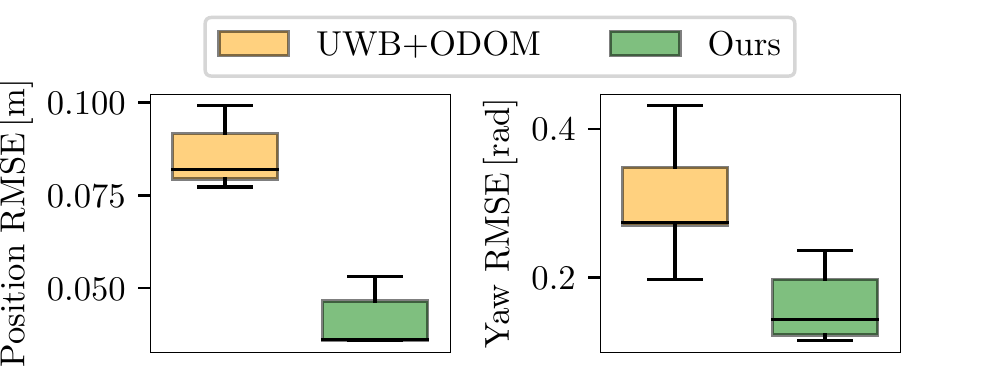}
	\caption{Position and orientation RMSE box plots from real experiments for our proposed method and the baseline (UWB+ODOM). The proposed method uses only range measurements whereas the baseline uses a combination of wheel encoder data and range measurements. The proposed method achieves better average position RMSE with lower deviation compared to the baseline. A similar trend can be observed in the orientation RMSE. Larger deviations of the baseline are due to wheel slippage.}
	\label{fig:rmse_plot_2d}
\end{figure}

\begin{table}[t]
	\setlength\extrarowheight{3.0pt}
	\caption{Average position and orientation RMSE of 2D trajectory estimation from real experiments.}
	\begin{tabularx}{\columnwidth}{@{}cCC@{}}
		\hline
		\hfil {Algorithm} & Average position RMSE\,[m] & Average orientation \hfil RMSE\,[rad]\\
		\hline
		UWB+ODOM & 0.086 & 0.303\\ 
		\hline
		Ours & \textbf{0.041} & \textbf{0.161} \\
		\hline
	\end{tabularx}
	\label{tab:2d_avg_rmse}
\end{table}

\begin{figure}[t]
	\hspace*{-1em}
	\includegraphics[scale=0.76]{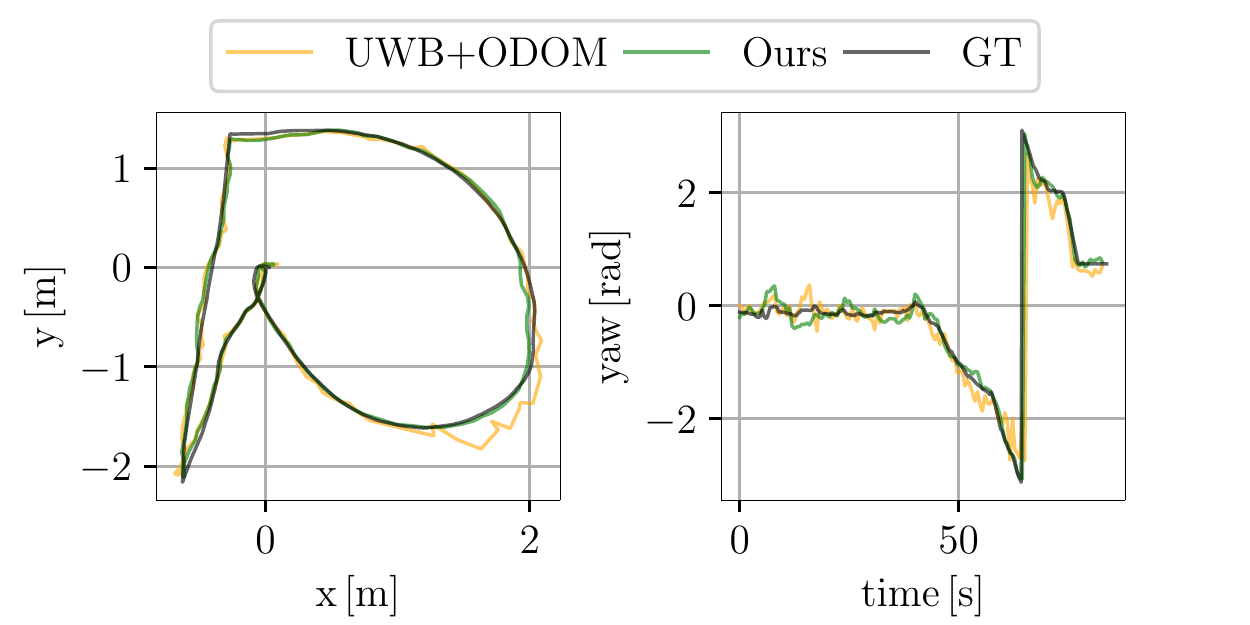}
	\caption{Trajectory estimation results from real experiments for our proposed method and the baseline (UWB+ODOM). The proposed method achieves similar tracking performance in position (left) and yaw angle (right) as the baseline without using wheel encoder data.}
	\label{fig:traj_plot_2d}
\end{figure}

\begin{figure}[t]
	\hspace*{0em}
	\includegraphics[scale=0.68]{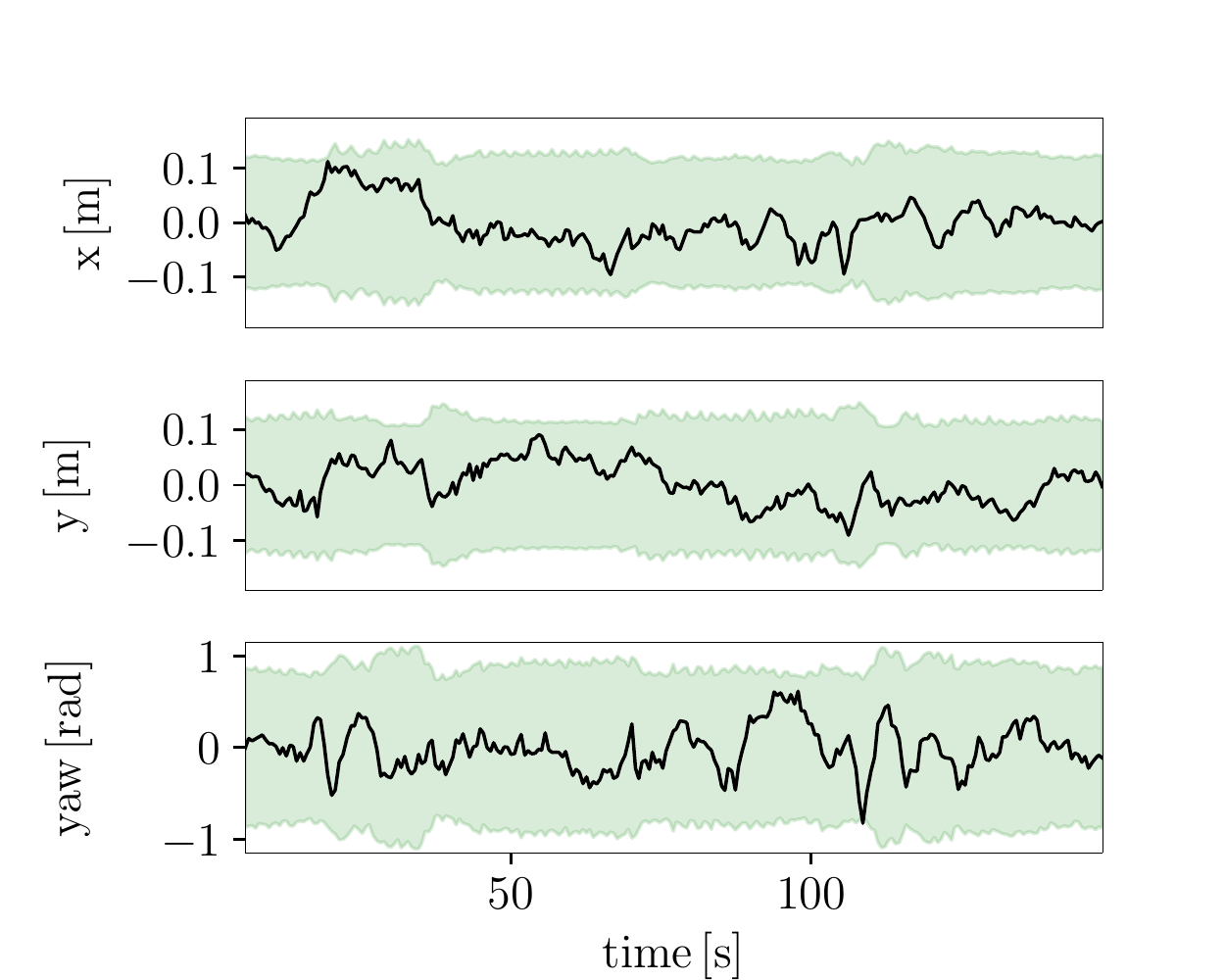}
	\caption{Error plots for position ($x$ and $y$) and yaw angle with the corresponding $3\sigma$ covariance envelopes for our proposed method from a real experiment. The estimation is unbiased and the estimated uncertainty bounds the observed error reliably. }
	\label{fig:err_plot_2d}
	\vspace*{-1em}
\end{figure}

\subsubsection*{2D Localization}

For 2D localization, we use a custom-built wheeled holonomic robot, shown in Figure \ref{fig:omniloc}, as our test platform. The robot is equipped with two UWB radios and three wheel encoders. The length of the lever arm is $\|\mathbf{p}^i_u\|_2 = 0.095\,\si{m}$. We compare our method with sensor fusion of range measurements and wheel odometry data \cite{magnago}. In this case, data from a single UWB radio is combined with velocity measurements from the wheel encoders using the same inference method as before. We refer to this baseline algorithm as UWB+ODOM.

We performed multiple experiments where the robot was driven manually along different trajectories in the test space and the sensor data was recorded onboard for offline evaluation of the two methods. UWB range data was processed to remove large outliers and constant biases for both algorithms.

Box plots of the position and the orientation RMSE for the two methods from 6 experiments are shown in Figure \ref{fig:rmse_plot_2d}. The proposed method achieves better average position RMSE with lower deviation compared to the baseline. A similar trend can be observed in the orientation RMSE. Larger deviations of the baseline method are due to frequent wheel slippage, which results in erroneous velocity measurements. The average RMSE values for the two methods are provided in Table \ref{tab:2d_avg_rmse}. Trajectory plots from one such experiment for the two methods are shown in Figure \ref{fig:traj_plot_2d}. The corresponding error plots with $3\sigma$ covariance bounds for our proposed method are shown in Figure \ref{fig:err_plot_2d}. The plots show that the estimated uncertainty bounds the observed error reliably.

The baseline achieves better tracking performance for trajectories involving rapid turns. This can be attributed to the lower update rate and sparsity of range measurements. Specifically, the wheel encoders provide linear and angular velocities at $20\,\si{Hz}$. In contrast, range measurements provide a single distance measurement at $17\,\si{Hz}$. With an increased range update rate, the proposed method should be able to track more aggressive trajectories. 

\begin{figure}[t]
	\centering
	\includegraphics[scale=0.35]{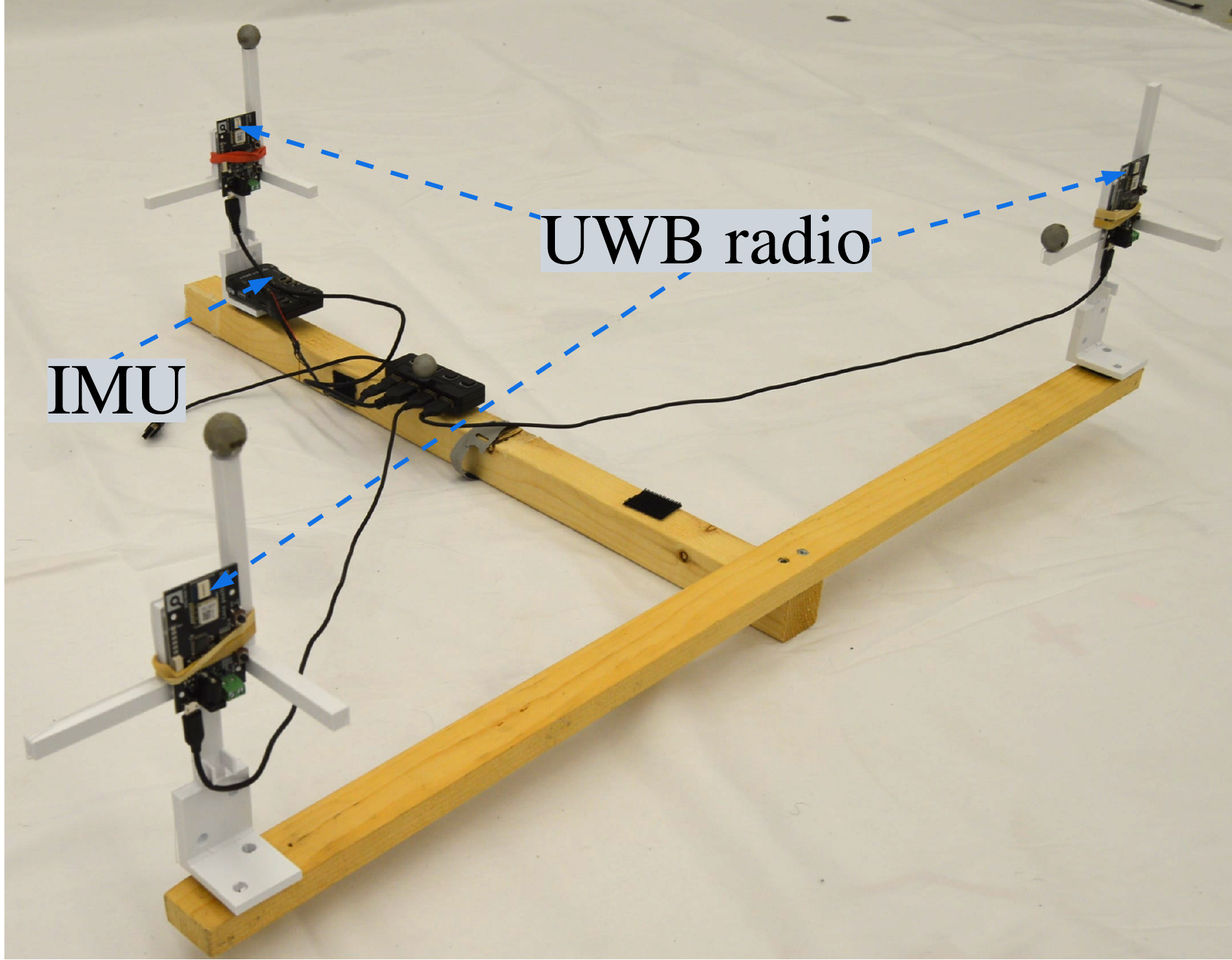}
	\caption{Our setup for 3D trajectory estimation is a sensor wand equipped with three ultrawideband (UWB) radios. The wand is also equipped with an inertial measurement unit (IMU), which is used for the baseline algorithm. The maximum length of the lever arm is $\| \mathbf{p}^i_u \|_2 = 0.72\,\si{m}$.}
	\label{fig:sensor_wand}
\end{figure}

\begin{figure*}[t]
	\hspace{-1em}
	\includegraphics[scale=0.7]{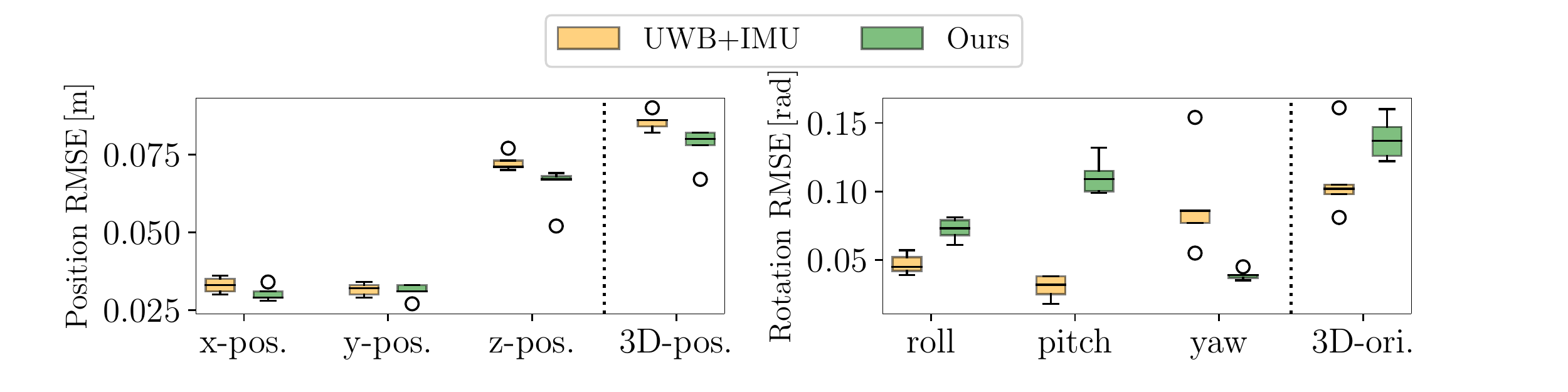}
	\caption{Individual and cumulative position and orientation RMSE from 3D trajectory estimation experiments for our proposed method and the baseline (UWB+IMU). The proposed method achieves lower yaw angle RMSE as the baseline method requires excitation for yaw to be estimated reliably. The proposed method has a higher roll and pitch angle RMSE due to the poor geometry of the test space along the z-axis and the range sensor separation on the sensor wand.}
	\label{fig:rmse_plot_3d}
	\vspace*{-1em}
\end{figure*}

\subsection{3D Localization}

Our test platform for 3D localization is a sensor wand equipped with 3 UWB radios and an IMU. The UWB radios are mounted in a noncollinear manner as shown in Figure \ref{fig:sensor_wand}. We compare our method to the tightly-coupled fusion of UWB and IMU data \cite{goudar2021}. Specifically, we combine measurements from a single UWB radio and an IMU using a fixed-lag smoother. We refer to this baseline as UWB+IMU.

\subsubsection*{Dynamic trajectories}
We performed multiple experiments where the sensor wand was moved manually along different trajectories to excite different axes of the IMU. The sensor data was recorded for offline evaluation of the two approaches.

\begin{figure*}[h]
%	\hspace{-1em}
	\includegraphics[scale=0.55]{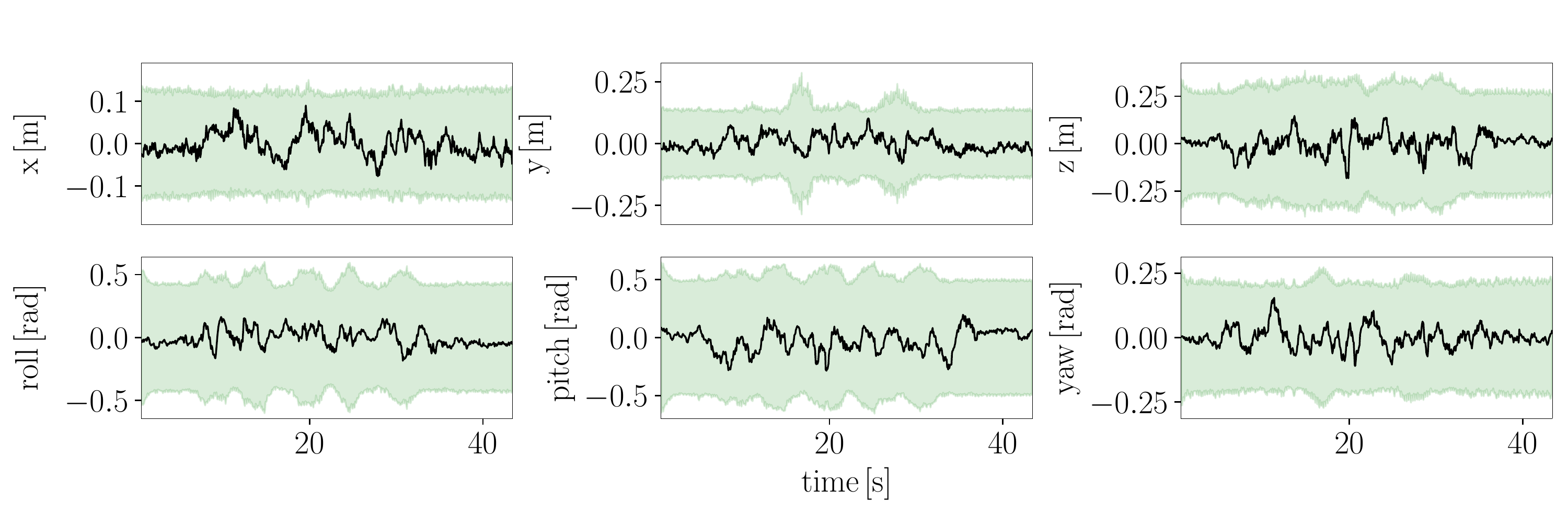}
	\caption{Error plots for the estimated position and  orientation (with the corresponding $3\sigma$ covariance envelopes) for our proposed method from a real experiment. The estimated uncertainty bounds the observed error. }
	\label{fig:error_plot_3d}
	\vspace*{-1em}
\end{figure*}

\begin{figure*}[h]
	%	\hspace{-1em}
	\includegraphics[scale=0.55]{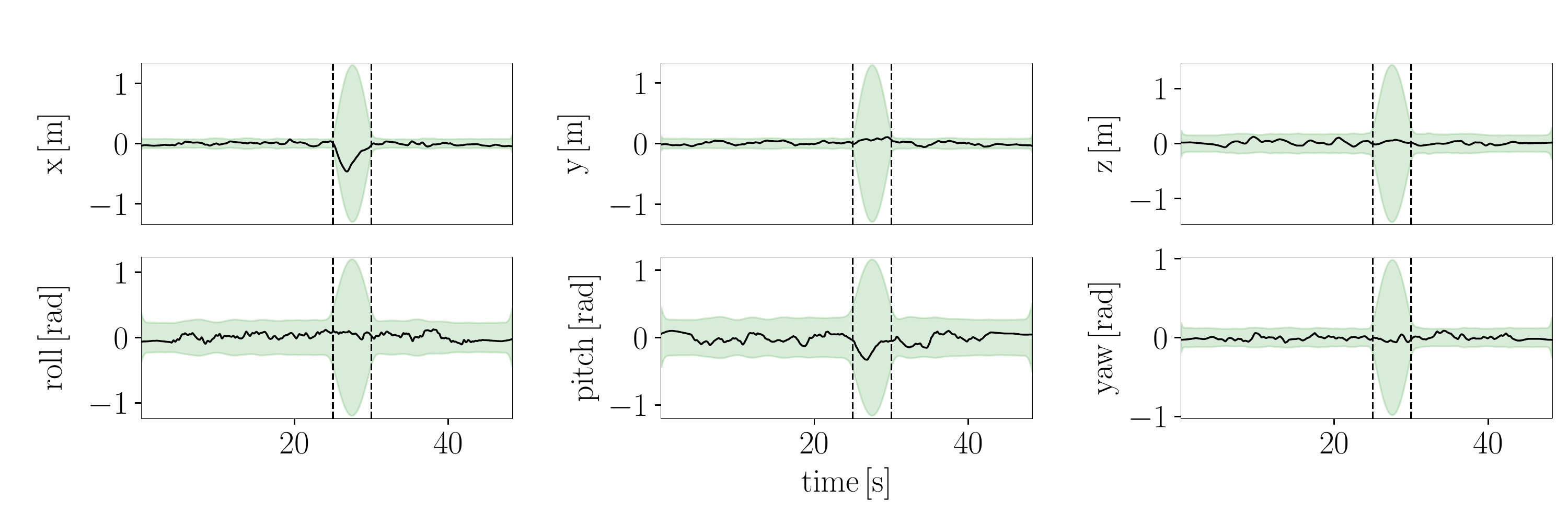}
	\caption{Error plots of the estimated trajectory (with the corresponding $3\sigma$ covariance envelopes) from real experiments with sensor dropout. Sensor data between $25\,\si{s}$ and $30\,\si{s}$ (indicated by dashed lines) is dropped to simulate sensor dropout. The proposed continuous-time approach is still able to estimate the trajectory reasonably well. The effect of sensor dropout is captured by the increased uncertainty about the state during the dropout period. }
	\label{fig:sens_dropout}
	\vspace*{-1em}
\end{figure*}

Box plots for axes-wise and cumulative position and orientation RMSE for the two methods from 5 experiments are shown in Figure \ref{fig:rmse_plot_3d}. The proposed method achieves a better position RMSE compared to the baseline. However, the baseline achieves a lower orientation RMSE compared to our method. This is expected as \textit{(i)} the IMU has angular rate measurements along \textit{each} body axis, and \textit{(ii)} the gravity vector is an accurate source of roll and pitch angle. However, the yaw angle RMSE of our proposed method is better compared to the baseline. This is because, in a tightly coupled UWB-IMU system, excitation of all the IMU axes is needed before the full pose becomes observable \cite{goudar2021}. In contrast, the proposed method does not require explicit excitation but does benefit from it. 
%For trajectories involving slower motion, the proposed method achieves lower orientation RMSE compared to baseline as seen from the maximum RMSE value of the baseline in Figure \ref{fig:rmse_plot_3d}.
The average RMSE values for the two methods are provided in Table \ref{tab:3d_avg_rmse}. Error plots along with the corresponding $3\sigma$ covariance envelopes from one of the experiments are shown in Figure \ref{fig:error_plot_3d}. The plots show that the observed error is bounded by the estimated uncertainty reliably. 

\begin{table}[t]
	\setlength\extrarowheight{2.5pt}
	\caption{Average position and orientation RMSE of 3D trajectory estimation from real experiments.}
	\begin{tabularx}{\columnwidth}{@{}cCC@{}}
		\hline
		Algorithm & Average position RMSE\,[m] & Average orientation RMSE\,[rad]\\
		\hline
		UWB+IMU & 0.086 & \textbf{0.109}\\ 
		\hline
		Ours & \textbf{0.078} & 0.138 \\
		\hline
	\end{tabularx}
	\label{tab:3d_avg_rmse}
\end{table}

\subsubsection*{Sensor dropout}
An advantage of the proposed continuous-time trajectory estimation method is that it can handle sensor dropouts. Specifically, in the absence of any sensor data, the motion prior is able to constrain the state. However, the baseline method is sensitive to IMU data dropout.  

For this setup, we use batch trajectory estimation instead of the FLS. We simulated sensor dropout by removing 5 seconds of measurements from the recorded data.  The proposed method estimates the trajectory reliably in this scenario as shown by the error plots in Figure \ref{fig:sens_dropout}. The effect of the sensor dropout is captured by the increased uncertainty around the state during the dropout period.

\section{Discussion}\label{sec:discussion}

The results show that reliable 2D and 3D pose estimation can be achieved with the proposed method, using only range measurements from multiple range sensors. However, the combination of UWB and IMU outperforms the proposed method for dynamic trajectories.  There are several factors that determine the efficacy of the proposed method in different settings. One of the factors is the bias in the measurements \cite{wenda}. In this work, the data was preprocessed to remove any constant biases. Any remnant biases can affect the estimation accuracy and hence need to be estimated online. Another factor that affects the estimation accuracy at high speeds is the update rate of the range sensors, which can be addressed using high-bandwidth high-rate sensors. A third factor that influences the estimation performance is the geometry of the installed anchors. Nonetheless, the results are promising considering that the precision of the UWB radios used in the experiment is $\pm 0.10\,\si{m}$. With higher-precision radio frequency technologies such as millimeter-wave radar, we expect to achieve higher accuracy.

\section{Conclusion} \label{sec:conclusion}

In this work, we presented a continuous-time trajectory estimation method for 2D and 3D pose estimation using only range measurements from multiple range sensors. Through simulation and real experiments, we showed that pose estimation can be done reliably using only range measurements. Additionally, the results show that the proposed method,  using off-the-shelf sensors, can achieve comparable performance and in some cases outperform conventional sensor fusion methods that require additional sensors.

There are many avenues for future work. One direction is to use different range-based measurement models such as time-difference-of-arrival \cite{dw1000}, which is more scalable. Another future direction is to evaluate different motion models such as the white-noise-on-jerk \cite{tangwnoj} motion model. Extension of the current method to continuous-time range-only multi-agent relative localization is another prospective future direction.

\bibliographystyle{unsrt}
\bibliography{main.bib}

\begin{thebibliography}{10}

\bibitem{kaplan}
Elliott~D. Kaplan.
\newblock {\em {Understanding GPS : principles and applications}}.
\newblock 1996.

\bibitem{raza}
Ali Raza, Lazar Lolic, Shahmir Akhter, and Michael Liut.
\newblock Comparing and evaluating indoor positioning techniques.
\newblock In {\em 2021 International Conference on Indoor Positioning and
  Indoor Navigation (IPIN)}, pages 1--8, 2021.

\bibitem{Blanco}
Jose-Luis Blanco, Javier Gonzalez, and Juan-Antonio Fernandez-Madrigal.
\newblock {A pure probabilistic approach to range-only SLAM}.
\newblock In {\em Proc. of the International Conference on Robotics and
  Automation (ICRA)}, pages 1436--1441, 2008.

\bibitem{hol2009}
Jeroen~D. Hol, Fred Dijkstra, Henk Luinge, and Thomas~B. Schon.
\newblock {Tightly coupled UWB/IMU pose estimation}.
\newblock In {\em International Conference on Ultra-Wideband}, pages 688--692.
  IEEE, 2009.

\bibitem{trawny2010}
Nikolas Trawny, Xun~S. Zhou, Ke~Zhou, and Stergios~I. Roumeliotis.
\newblock {Interrobot Transformations in 3-D}.
\newblock {\em IEEE Transactions on Robotics}, 26(2):226--243, 2010.

\bibitem{goudar2021}
Abhishek Goudar and Angela~P Schoellig.
\newblock {Online Spatio-temporal Calibration of Tightly-coupled
  Ultrawideband-aided Inertial Localization}.
\newblock pages 1161--1168. IEEE, 2021.

\bibitem{yavari2014}
Mohammadreza Yavari and Bradford~G Nickerson.
\newblock {Ultra wideband wireless positioning systems}.
\newblock {\em Technical Report TR14-230}, 506, 2014.

\bibitem{Djugash}
Joseph Djugash.
\newblock {\em Geolocation with Range: Robustness, Efficiency and Scalability}.
\newblock PhD thesis, Carnegie Mellon University, Pittsburgh, PA, November
  2010.

\bibitem{zhouwheel}
Boli Zhou, Hongbin Fang, and Jian Xu.
\newblock Uwb-imu-odometer fusion localization scheme: Observability analysis
  and experiments.
\newblock {\em IEEE Sensors Journal}, 23(3):2550--2564, 2023.

\bibitem{Mueller2015}
Mark~W. Mueller, Michael Hamer, and Raffaello D'Andrea.
\newblock {Fusing ultra-wideband range measurements with accelerometers and
  rate gyroscopes for quadrocopter state estimation}.
\newblock volume 2015-June, pages 1730--1736. IEEE, 2015.

\bibitem{Hoeller2017}
David Hoeller, Anton Ledergerber, Michael Hamer, and Raffaello D'Andrea.
\newblock {Augmenting Ultra-Wideband Localization with Computer Vision for
  Accurate Flight}.
\newblock {\em IFAC-PapersOnLine}, 50(1):12734--12740, 2017.

\bibitem{Nguyen2021}
Thien~Hoang Nguyen, Thien-Minh Nguyen, and Lihua Xie.
\newblock {Range-focused Fusion of Camera-IMU-UWB for Accurate and
  Drift-reduced Localization}.
\newblock {\em IEEE Robotics and Automation Letters}, 6(2):1--1, 2021.

\bibitem{cioffi2020tightly}
Giovanni Cioffi and Davide Scaramuzza.
\newblock {Tightly-coupled fusion of global positional measurements in
  optimization-based visual-inertial odometry}.
\newblock In {\em Proc. of the International Conference on Intelligent Robots
  and Systems (IROS)}, pages 5089--5095. IEEE, 2020.

\bibitem{splineimu}
Kailai Li, Ziyu Cao, and Uwe~D. Hanebeck.
\newblock {Continuous-Time Ultra-Wideband-Inertial Fusion}.
\newblock {\em arXiv preprint arXiv:2301.09033}, January 2023.

\bibitem{kiki2019}
Michalina Pacholska, Frederike D{\"u}mbgen, and Adam Scholefield.
\newblock {Relax and Recover: Guaranteed Range-Only Continuous Localization}.
\newblock {\em IEEE Robotics and Automation Letters}, 5:2248--2255, 2019.

\bibitem{Barfoot2014a}
Timothy~D. Barfoot, Chi~Hay Tong, and Simo S{\"{a}}rkk{\"{a}}.
\newblock {Batch Continuous-Time Trajectory Estimation as Exactly Sparse
  Gaussian Process Regression}.
\newblock {\em Robotics: Science and Systems}, 2014.

\bibitem{kiki2022}
Frederike D{\"u}mbgen, Connor~T. Holmes, and Tim~D. Barfoot.
\newblock {Safe and Smooth: Certified Continuous-Time Range-Only Localization}.
\newblock {\em IEEE Robotics and Automation Letters}, 8:1117--1124, 2022.

\bibitem{fishberg}
Andrew Fishberg and Jonathan~P How.
\newblock {Multi-Agent relative pose estimation with UWB and constrained
  communications}.
\newblock In {\em Proc. of the International Conference on Intelligent Robots
  and Systems (IROS)}, pages 778--785. IEEE, 2022.

\bibitem{arunsmethod}
K.~S. Arun, T.~S. Huang, and S.~D. Blostein.
\newblock {Least-Squares Fitting of Two 3-D Point Sets}.
\newblock {\em IEEE Transactions on Pattern Analysis and Machine Intelligence},
  PAMI-9(5):698--700, 1987.

\bibitem{Anderson2015}
Sean Anderson and Timothy~D. Barfoot.
\newblock {Full STEAM ahead: Exactly sparse Gaussian process regression for
  batch continuous-time trajectory estimation on SE(3)}.
\newblock volume 2015-Decem, pages 157--164. IEEE, 2015.

\bibitem{Barfoot2023}
Timothy~D. Barfoot.
\newblock {\em {State estimation for robotics}}.
\newblock Cambridge University Press, {Second} edition, 2023.

\bibitem{chirikjian}
Gregory~S Chirikjian.
\newblock {\em Stochastic models, information theory, and Lie groups, volume 2:
  Analytic methods and modern applications}.
\newblock Springer Science \& Business Media, 2011.

\bibitem{gtsam}
Frank~Dellaert et~al.
\newblock {borglab/gtsam}.
\newblock \url{https://doi.org/10.5281/zenodo.5794541}, May 2022.

\bibitem{dw1000}
{Decawave DW1000 UWB transceiver}.
\newblock \url{https://fcc.report/FCC-ID/2AAXVTNTMOD1/2787937.pdf}.
\newblock Accessed: 2022-12-10.

\bibitem{magnago}
Valerio Magnago, Pablo Corbal{\'a}n, Gian~Pietro Picco, Luigi Palopoli, and
  Daniele Fontanelli.
\newblock {Robot localization via odometry-assisted ultra-wideband ranging with
  stochastic guarantees}.
\newblock In {\em Proc. of the International Conference on Intelligent Robots
  and Systems (IROS)}, pages 1607--1613. IEEE, 2019.

\bibitem{wenda}
Wenda Zhao, Abhishek Goudar, Jacopo Panerati, and Angela~P. Schoellig.
\newblock Learning-based bias correction for ultra-wideband localization of
  resource-constrained mobile robots.
\newblock {\em arXiv preprint arXiv:2003.09371}, 2020.

\bibitem{tangwnoj}
Tim~Yuqing Tang, David~Juny Yoon, and Timothy~D. Barfoot.
\newblock {A White-Noise-on-Jerk Motion Prior for Continuous-Time Trajectory
  Estimation on SE(3)}.
\newblock {\em IEEE Robotics and Automation Letters}, 4(2):594--601, 2019.

\end{thebibliography}


\begin{thebibliography}{10}

\bibitem{kaplan}
Elliott~D. Kaplan.
\newblock {\em {Understanding GPS : principles and applications}}.
\newblock 1996.

\bibitem{raza}
Ali Raza, Lazar Lolic, Shahmir Akhter, and Michael Liut.
\newblock {Comparing and Evaluating Indoor Positioning Techniques}.
\newblock pages 1--8, 11 2021.

\bibitem{Blanco}
Jose-Luis Blanco, Javier Gonzalez, and Juan-Antonio Fernandez-Madrigal.
\newblock {A pure probabilistic approach to range-only SLAM}.
\newblock In {\em 2008 IEEE International Conference on Robotics and
  Automation}, pages 1436--1441, 2008.

\bibitem{hol2009}
Jeroen~D. Hol, Fred Dijkstra, Henk Luinge, and Thomas~B. Schon.
\newblock {Tightly coupled UWB/IMU pose estimation}.
\newblock In {\em 2009 IEEE International Conference on Ultra-Wideband}, pages
  688--692, 2009.

\bibitem{trawny2010}
Nikolas Trawny, Xun~S. Zhou, Ke~Zhou, and Stergios~I. Roumeliotis.
\newblock {Interrobot Transformations in 3-D}.
\newblock {\em IEEE Transactions on Robotics}, 26(2):226--243, 2010.

\bibitem{goudar2021}
Abhishek Goudar and Angela~P Schoellig.
\newblock {Online Spatio-temporal Calibration of Tightly-coupled
  Ultrawideband-aided Inertial Localization}.
\newblock In {\em Proc. of the International Conference on Intelligent Robots
  and Systems}, pages 1161--1168. IEEE, 2021.

\bibitem{yavari2014}
Mohammadreza Yavari and Bradford~G Nickerson.
\newblock {Ultra wideband wireless positioning systems}.
\newblock {\em Technical Report TR14-230}, 506, 2014.

\bibitem{Djugash}
PhD thesis.

\bibitem{zhouwheel}
Boli Zhou, Hongbin Fang, and Jian Xu.
\newblock {UWB-IMU-Odometer Fusion Localization Scheme: Observability Analysis
  and Experiments}.
\newblock {\em IEEE Sensors Journal}, 23(3):2550--2564, 2023.

\bibitem{Mueller2015}
Mark~W. Mueller, Michael Hamer, and Raffaello D'Andrea.
\newblock {Fusing ultra-wideband range measurements with accelerometers and
  rate gyroscopes for quadrocopter state estimation}.
\newblock volume 2015-June, pages 1730--1736. IEEE, 2015.

\bibitem{Hoeller2017}
David Hoeller, Anton Ledergerber, Michael Hamer, and Raffaello D'Andrea.
\newblock {Augmenting Ultra-Wideband Localization with Computer Vision for
  Accurate Flight}.
\newblock {\em IFAC-PapersOnLine}, 50(1):12734--12740, 2017.

\bibitem{Nguyen2021}
Thien~Hoang Nguyen, Thien-Minh Nguyen, and Lihua Xie.
\newblock {Range-focused Fusion of Camera-IMU-UWB for Accurate and
  Drift-reduced Localization}.
\newblock {\em IEEE Robotics and Automation Letters}, 6(2):1--1, 2021.

\bibitem{splineimu}
Kailai Li, Ziyu Cao, and Uwe~D. Hanebeck.
\newblock Continuous-time ultra-wideband-inertial fusion.
\newblock {\em arXiv preprint arXiv:2301.09033}, January 2023.

\bibitem{kiki2022}
Frederike D{\"u}mbgen, Connor~T. Holmes, and Tim~D. Barfoot.
\newblock {Safe and Smooth: Certified Continuous-Time Range-Only Localization}.
\newblock {\em IEEE Robotics and Automation Letters}, 8:1117--1124, 2022.

\bibitem{Barfoot2014a}
Timothy~D. Barfoot, Chi~Hay Tong, and Simo S{\"{a}}rkk{\"{a}}.
\newblock {Batch Continuous-Time Trajectory Estimation as Exactly Sparse
  Gaussian Process Regression}.
\newblock {\em Robotics: Science and Systems}, 2014.

\bibitem{fishberg}
Andrew Fishberg and Jonathan~P. How.
\newblock {Multi-Agent Relative Pose Estimation with UWB and Constrained
  Communications}.
\newblock In {\em 2022 IEEE/RSJ International Conference on Intelligent Robots
  and Systems (IROS)}, pages 778--785, 2022.

\bibitem{arunsmethod}
K.~S. Arun, T.~S. Huang, and S.~D. Blostein.
\newblock Least-squares fitting of two 3-d point sets.
\newblock {\em IEEE Transactions on Pattern Analysis and Machine Intelligence},
  PAMI-9(5):698--700, 1987.

\bibitem{Anderson2015}
Sean Anderson and Timothy~D. Barfoot.
\newblock {Full STEAM ahead: Exactly sparse Gaussian process regression for
  batch continuous-time trajectory estimation on SE(3)}.
\newblock {\em Proc. of the IEEE International Conference on Intelligent Robots
  and Systems}, 2015-Decem(3):157--164, 2015.

\bibitem{Barfoot2023}
Timothy~D. Barfoot.
\newblock {State estimation for robotics}.
\newblock {Second edition}, 2023.

\bibitem{chirikjian2011}
Gregory~S Chirikjian.
\newblock {\em {Stochastic models, information theory, and Lie groups, volume
  2: Analytic methods and modern applications}}, volume~2.
\newblock Springer Science \& Business Media, 2011.

\bibitem{gtsam}
Frank~Dellaert et~al.
\newblock {borglab/gtsam}, May 2022.

\bibitem{dw1000}
{Decawave DW1000 UWB transceiver}.
\newblock Accessed: 2022-12-10.

\bibitem{imupreint}
Christian Forster, Luca Carlone, Frank Dellaert, and Davide Scaramuzza.
\newblock {On-Manifold Preintegration for Real-Time Visual--Inertial Odometry}.
\newblock {\em IEEE Transactions on Robotics}, 33(1):1--21, 2017.

\bibitem{wong2020}
Jeremy~N. Wong, David~J. Yoon, Angela~P. Schoellig, and Timothy~D. Barfoot.
\newblock {A Data-Driven Motion Prior for Continuous-Time Trajectory Estimation
  on SE(3)}.
\newblock {\em IEEE Robotics and Automation Letters}, 5(2):1429--1436, 2020.

\end{thebibliography}

% that's all folks
\end{document}